\pdfoutput=1

\documentclass[11pt]{article}

\usepackage[final]{acl}

\usepackage{times}
\usepackage{latexsym}

\usepackage[T1]{fontenc}

\usepackage[utf8]{inputenc}

\usepackage{subfigure}

\usepackage{microtype}

\usepackage{inconsolata}

\usepackage{graphicx}

\usepackage{tabularray}
\UseTblrLibrary{booktabs}

\title{\textit{Dialetto, ma Quanto Dialetto?} \\Transcribing and Evaluating Dialects on a Continuum} 

\author{Ryan Soh-Eun Shim\textsuperscript{\normalfont 1, 2} \and
         Barbara Plank\textsuperscript{\normalfont 1, 2}\\
  \textrm{\textsuperscript{1}}MaiNLP, Center for Information and Language Processing, LMU Munich, Germany \\
  \textrm{\textsuperscript{2}}Munich Center for Machine Learning (MCML), Munich, Germany \\
{\tt s.shim@lmu.de} \hspace{2em} {\tt b.plank@lmu.de}}

\begin{document}
\maketitle
\begin{abstract}
There is increasing interest in looking at dialects in NLP. However, most work to date still treats dialects as discrete categories. For instance, evaluative work in variation-oriented NLP for English often works with Indian English or African-American Venacular English as homogeneous categories, yet even within one variety there is substantial variation. We examine within-dialect variation and show that performance critically varies within categories. We measure speech-to-text performance on Italian dialects, and empirically observe a geographical performance disparity. This disparity correlates substantially (-0.5) with linguistic similarity to the highest performing dialect variety. We cross-examine our results against dialectometry methods, and interpret the performance disparity to be due to a bias towards dialects that are more similar to the standard variety in the speech-to-text model examined. We additionally leverage geostatistical methods to predict zero-shot performance at unseen sites, and find the incorporation of geographical information to substantially improve prediction performance, indicating there to be geographical structure in the performance distribution.
\end{abstract}

\section{Introduction}
An increasing body of work in Natural Language Processing (NLP) has called attention to the disparity in research focus between high-resource, standardized linguistic varieties and empirical linguistic variation \citep{plank2016nonstandardornoncanonicallanguage, kantharuban-etal-2023-quantifying, chang2024self}. While there are many types of variation (e.g.\ genre, register), \emph{dialect} variation has emerged as a particular point of focus, with increasing availability of evaluative benchmarks \citep{faisal-etal-2024-dialectbench, ziems-etal-2023-multi}, dialect-specific datasets \citep{doganschönberger2021swissdialparallelmultidialectalcorpus, blaschke-etal-2024-maibaam}, and methodological contributions \citep{blaschke-etal-2023-manipulating, demszky-etal-2021-learning} towards dialect-robust models \citep{zampieri2020natural}. 

A considerable amount of work conceptualizes dialects solely as \emph{discrete} linguistic categories that stand side-by-side with the standard variety (e.g.\ African-American Vernacular English vs.\ mainstream American English) \citep{faisal-etal-2024-dialectbench, ziems-etal-2023-multi}. However, prior work in dialectology has noted that dialect relations often stand in a \emph{continuum}, where similarity between varieties slowly decreases the further away they are from a given geographical site, rather than being a sharp transition \citep{heeringa2001dialect}. For dialect NLP, this means that a purely discrete conceptualization of linguistic categories not only largely overlooks the dialect continuum, but also leaves 
important \textit{evaluation gaps},  which can even lead to social harm. 
This is because gradient variation within the category may not be evenly described \citep{jones-2015-aave-variation, labov2012dialect}, and lesser known transitional varieties \citep{jeszenszky2018gradient} between the linguistic categories examined may be left out of evaluative benchmarks. 

\begin{figure*}
  \includegraphics[width=\textwidth]{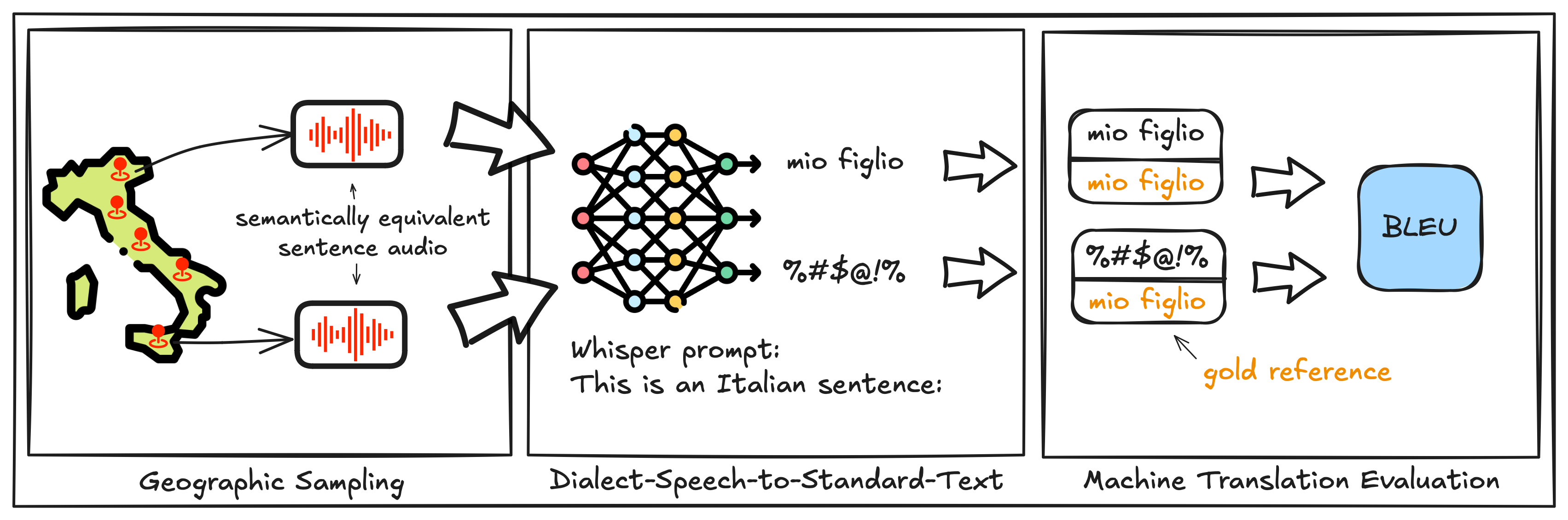}
  \caption{Procedure through which we obtain our speech-to-text scores. Semantically-equivalent sentence audios are sampled across geographically-balanced dialects, which are then transcribed to the standard variety with speech-to-text models, and evaluated with machine translation evaluation metrics.} 
\end{figure*}

Concretely, this paper addresses two research questions:

\begin{enumerate}
    \item[] RQ1: Is the distribution of speech-to-text performance on dialect speech a \textit{geographically autocorrelated} variable? 
    
    \item[] RQ2: To what extent is the distribution predictable by way of \textit{phonetic similarity} to the best performing variety? 
\end{enumerate}

Our first question (RQ1) is motivated by prior work in the geosciences, which has consistently relied on the insight that ``near things are more related than distant things''---a principle known as Tobler's first law of geography \citep{tobler1970computer}---for interpolation of missing values \citep{matheron1963principles}. On the other hand, our second question (RQ2) is motivated by the possibility of pretraining data containing differing amounts of dialect speech, which is likely to serve as a confounding variable to the correlation between phonetic similarity and speech-to-text performance.

To answer the two questions, we perform a fine-grained investigation of such a regional gap in speech-to-text models to gain insights on cross-lingual transfer performance by drawing upon geostatistical and dialectometric techniques. In contrast with prior work which evaluates performance disparity on a discrete basis (e.g.\ Chilean Spanish, Argentinian Spanish) \citep{kantharuban-etal-2023-quantifying}, we place dialects on a continuum using a large-scale \emph{geotagged} dataset of related Italian dialects, where we conceptualize dialect relations to stand in a continuum \citep{heeringa2001dialect}. As such, we perform zero-shot speech-to-text on semantically-equivalent speech samples across geographically-contiguous dialect sites, where in line with \citet{kantharuban-etal-2023-quantifying}, we find evidence of zero-shot performance of speech-to-text models on dialects correlating with similarity to the standard variety.\footnote{Our code is available here: \url{https://github.com/mainlp/dialetto}} Our contributions are as follows:

\begin{enumerate}
    \item \textbf{Categorical to Continuous Conceptualization}:
    Following established work in linguistic dialectometry \citep{heeringa2001dialect}, we conceptualize dialect relations as a continuum. This allows us to visualize in fine-grained detail regional performance gaps, which we find to correlate strongly with linguistic similarity to the highest performing variety, corroborating prior evaluative work done on a categorical basis \citep{kantharuban-etal-2023-quantifying}.
    \item \textbf{Geostatistics for Dialect NLP}: We perform a dialect-level examination of zero-shot performance prediction, and leverage geostatistical techniques for interpolating performance at held-out sites. We find the incorporation of geographical information to lead to a robust increase in performance prediction.
\end{enumerate}

\section{Dataset}
\subsection{Italian Dialect Dataset}
We conduct our study on Vivaldi \citep{tosques2013vivaio}, a geo-tagged parallel corpus of spoken Italian dialect varieties in audio form. The corpus contains data at different levels of linguistic units (e.g. word, sentence, discourse). The data is collected across 293 sites in Italy, and is divided into the categories of phonetic, lexical, morphological, syntactic, and discourse level data. Our criterion for using semantically-equivalent data across dialect sites is motivated by the fact that there are syntactic, lexical, and phonetic differences between dialects, which serves as a more realistic basis for evaluating zero-shot performance on dialects. As such, we leverage 15 sentence-level recordings and 20 word-level recordings per site in our experiments. To ensure a fair comparison between sites and because of the rich linguistic variety observed in Italy~\cite{ramponi-2024-language}, we further filter for only dialect sites that fall under the Italic language branch according to the metadata, thereby removing data from Bavarian, Greek, Occitan, among others. This results in 223 sites. \autoref{table:datasets} summarizes the statistical information of our data. 

In view of the scale of variation in Italian dialects, where differences between groups may warrant the status of individual languages, we also evaluate our results on only the Tuscan subgroup, which qualitative work establishes as bearing the most similarity to standard Italian \citep{wieling2014lexical}. 

\begin{table}
\small
\centering
\SetTblrInner{rowsep=0pt}
\begin{tblr}{ccc}
\toprule
\textbf{\# dialects used} & \textbf{\# words used} & \textbf{\# sentences used} \\
\midrule
223 & 20 & 15 \\
\bottomrule
\end{tblr}
\caption{Statistics on the subset used of Vivaldi \citep{tosques2013vivaio}, an Italian dialect corpus.}
\label{table:datasets}
\end{table}


\section{Methodology}
\subsection{Speech-to-Text}
For our speech-to-text model, we employ Whisper \citep{radford2023robust}, a family of encoder-decoder speech-to-text models. Our experiment utilizes Whisper-large-v3, which is trained on 1 million hours of weakly-labeled and 4 million hours of pseudo-labeled audio data. The training regime for Whisper is both multilingual and multi-task, where samples are either asked to be transcribed into the original language, or to be translated into English. This is achieved by way of special tokens (e.g. <lang>, <translate>, <transcribe>). In addition, the data format for long-form transcriptions includes a token <prev> to denote the previous context during training. At inference time, the space of this prior context can be used to achieve prompting, where the speech-to-text output would be conditioned on this prior context. We take advantage of this prior context to prompt the model to transcribe the speech in standard language (e.g. "\textit{Questa è una frase italiana: }"; English translation: "this is an Italian sentence: "). This is necessary due to the spoken nature of dialect varieties in Italy, where there is often no widely used written variety that corresponds with what is spoken, and speakers of such varieties would write often only in the standard variety. 

\begin{figure*}[htp]
  \centering
  \includegraphics[width=0.7\linewidth]{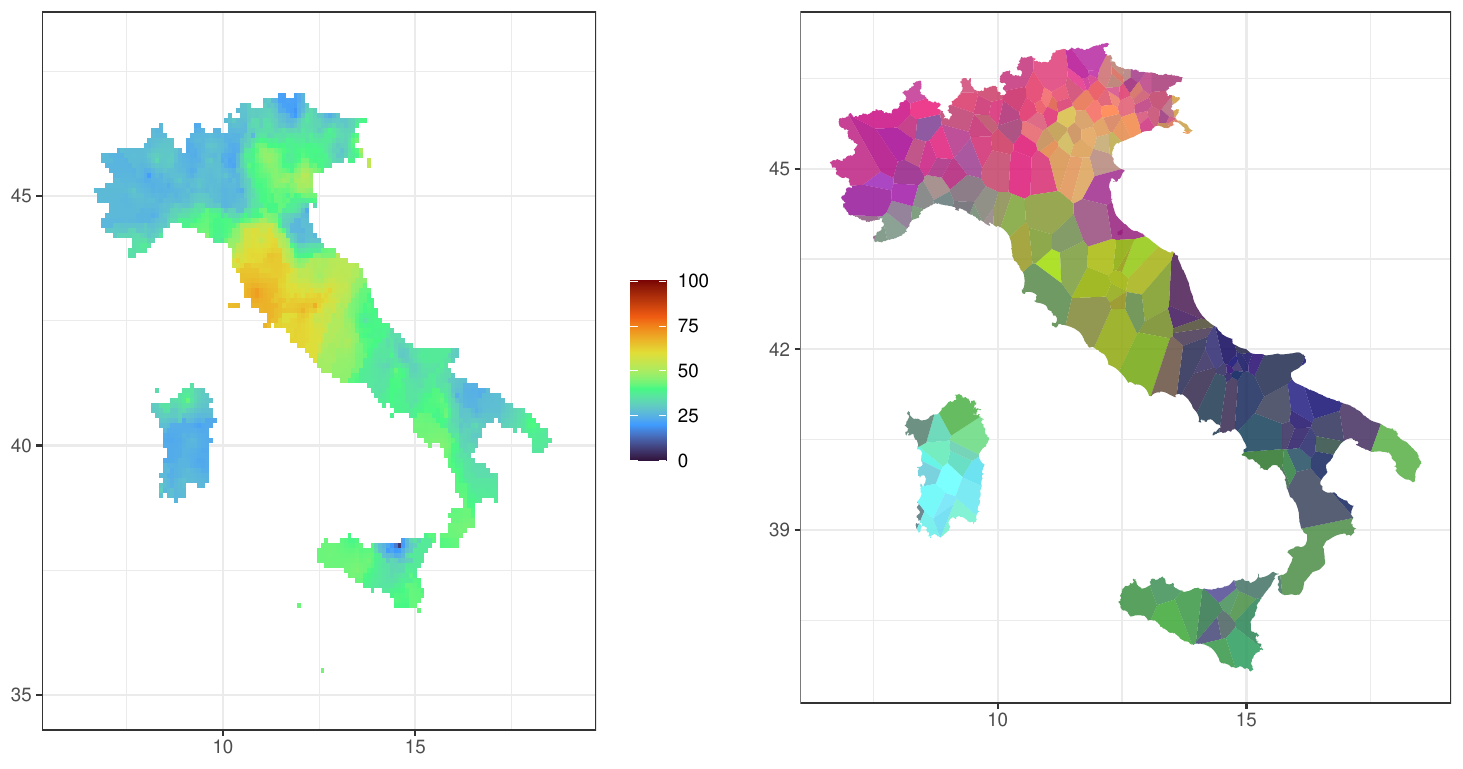}
  \caption{Left plot: chrF2 of zero-shot speech-to-text on Italian for Whisper-large-v3 interpolated with inverse distance weighting (left). Red-yellow area is Tuscany, from which standard Italian comes. Right plot: MDS-based dialectometry visualization.}
  \label{fig:maps}
\end{figure*}

\subsection{Evaluation}
\citet{dolev-etal-2024-whisper} report Whisper as a viable system for speech-based dialect-to-standard speech-to-text for Swiss German when transcribed to standard German text and evaluated with BLEU \citep{papineni-etal-2002-bleu}. Given a similar mismatch between input and output for dialect speech to standard text for our Italian dialect data, we follow \citet{dolev-etal-2024-whisper} in evaluating our dialect speech-to-text output with standard machine translation evaluation metrics. We employ BLEU~\citep{papineni-etal-2002-bleu} and chrF \citep{popovic-2015-chrf}\footnote{We employ SacreBLEU \citep{post-2018-call}.} both of which are based on n-gram overlap. Due to the expected mismatch between Italian dialect speech and standard Italian text, we increase the number of gold references to allow for more opportunity for alternative yet valid phrasings to be counted as correct. We follow prior work in expanding the number of gold references by way of a LLM-based paraphrasing approach \citep{tang-etal-2024-metrics, zeng-etal-2024-towards}. \autoref{table:llm_expand} gives an example of the original and generated gold references for standard Italian. In our experiments, we generate 10 additional references per item,\footnote{We use Meta-Llama-3.1-70B-Instruct: \url{https://huggingface.co/meta-llama/Llama-3.1-70B-Instruct}} in addition to the original gold standard. Note that we do not employ standard ASR metrics such as word error rate or character error rate, as we do not expect the speech and the text to align well for every variety, due to the non-written nature of non-standard varieties.

\begin{table}
\resizebox{\columnwidth}{!}{
\centering
\SetTblrInner{rowsep=0pt}
\begin{tblr}{lll}
\toprule
& \textbf{Original} & \textbf{Generated} \\
\midrule
\midrule
(1)  & Si munge due volte al giorno & Si munge due volte ogni giorno \\
\midrule
\textbf{EN}: & \textit{One milks two times per day} & \textit{One milks two times every day} \\
\midrule
\midrule
(2) & Domani tornerò a casa & Domani ritornerò a casa mia \\
\midrule
\textbf{EN}: & \textit{I will go home tomorrow} & \textit{I will return to my home tomorrow} \\
\bottomrule
\end{tblr}
} 
\caption{Examples of LLM-generated additional reference translations.}
\label{table:llm_expand}
\end{table}


\subsection{Geostatistical Analysis}
In this section, we introduce geostatistical \citep{matheron1963principles, cressie1989geostatistics} methods for the interpolation of speech-to-text performance at unseen sites, where geostatistics refers to a family of statistical techniques designed to model spatial data. Our introduction of such methods is driven by two considerations: to understand whether the geographical distribution of model performance is indeed sufficiently autocorrelated for such interpolation to work, and for the practical concern that data collected for dialects may be more sparsely distributed than desired due to difficulties in collection, thus raising the need for interpolation.

Prior work has found geographical proximity between pivot and target language to be an important predictor of cross-lingual transfer \citep{ahuja-etal-2022-multi, samardzic-etal-2022-language, lin-etal-2019-choosing}. The varieties examined in \citet{ahuja-etal-2022-multi} cover the span of \textit{languages}, where the performance is argued to be due to overlap in typological and vocabulary overlap. We propose that for varieties \textit{within} the same language, the geographical signal is arguably even stronger due to higher geographical proximity and fewer confounding factors \citep{shim-etal-2024-phonotactic, JeszenszkyStoeckleGlaserWeibel2017}, such that the signal can be helpful in predicting zero-shot performance for varieties within a given language. To verify this claim, we leverage geostatistical techniques to predict the zero-shot performance of Whisper on unseen held-out sites. We employ three geostatistical interpolation methods in our experiments:\footnote{We use implementations in the R package gstat \citep{gstat1,gstat2}.} nearest neighbor interpolation (NN) \citep{sibson1981brief}, inverse distance weighting (IDW) \citep{shepard1968two}, and kriging \citep{oliver1990kriging}. While NN and IDW do not make assumptions on the geographical distribution of the data, the use of kriging makes the assumption of stationarity, which is the assumption that the mean and variance are constant across space. Given that the variable we model is speech-to-text performance on dialects, its geographical distribution arguably may violate the assumption of stationarity, depending on how well the model generalizes to the dialect varieties in different regions. Given such a potential for non-stationarity in the data, we expect this to require modelling in order to better satisfy the assumption. We detail our treatment of this in Section \ref{RK}.

In our experiments, we perform an 80/10/10 split for training, validation, and test data, where we tune hyperparameters by way of a grid search on the validation set, and report the root mean square error (RMSE) on the test set. In addition, we measure the impact of training data size across the geostatistical methods examined, where we sample the training data from a percentage range of 0.1 to 1.0. For each point in the percentage range, we repeat the sampling procedure 100 times and take the mean of the interpolation RMSE across the 100 runs, in order to ensure that the performance reported is representative of the data. \autoref{table:geo_rmse} summarizes our results; \autoref{fig:geocurve} shows the effect of training data size.

\subsubsection{Baseline}
To verify the extent to which increasing levels of geospatial information is helpful for prediction, we employ nearest neighbor interpolation \citep{sibson1981brief} as a baseline, where the predicted value of a sample is taken to be identical to its geographically nearest neighbor in the training data.

\subsubsection{Inverse Distance Weighting}
Inverse distance weighting (IDW) \citep{shepard1968two} is a geostatistical interpolation method, where the estimated value at a target point \( \hat{v}(x_0) \) is computed as a weighted average of the known values from surrounding data points. The weights are inversely proportional to the distance from the target point, with closer points having more influence. Formally, given a set of known points \( x_1, x_2, \dots, x_n \) with corresponding values \( v(x_1), v(x_2), \dots, v(x_n) \), the interpolated value at \( x_0 \) is defined as:

\[
\hat{v}(x_0) = \frac{\sum_{i=1}^{n} w_i(x_0) v(x_i)}{\sum_{i=1}^{n} w_i(x_0)},
\]

where the weights \( w_i(x_0) \) are given by:

\[
w_i(x_0) = \frac{1}{d(x_0, x_i)^p}
\]

with \( d(x_0, x_i) \) representing the Euclidean distance between \( x_0 \) and \( x_i \), and \( p \) being a positive power parameter that controls the influence of the distance. In our experiments, we perform a grid search and evaluate on the validation data to determine the hyperparameters used on the test data for each run.

\subsubsection{Variogram}

Our method of kriging \citep{oliver1990kriging} relies on the concept of a variogram, where a variogram \citep{cressie1985fitting} is a method in geostatistics that is used to quantify the degree of spatial autocorrelation between data points. In contrast to IDW, which assumes a deterministic decrease in similarity as distance increases, a variogram provides a probabilistic approach to measuring how spatial correlation between values changes with increasing separation distance, allowing for a more data-driven approach towards deriving the weights of the values used to interpolate the value at an unknown site. Formally, the variogram \( \gamma(h) \) is defined as:

\[
\gamma(h) = \frac{1}{2N(h)} \sum_{i=1}^{N(h)} \left[ Z(x_i) - Z(x_i + h) \right]^2
\]

where \( Z(x_i) \) is the observed value of the random variable \( Z \) at location \( x_i \); \( h \) is the lag distance, representing the separation distance between two locations; \( Z(x_i + h) \) is the value of \( Z \) at location \( x_i + h \); and \( N(h) \) is the number of data point pairs separated by distance \( h \).

The function \( \gamma(h) \) estimates the spatial variance as a function of distance, with larger values of \( h \) indicating weaker correlation between points. A variogram is typically visualized by plotting \( \gamma(h) \) against \( h \), which allows for a visual interpretation of the y-intercept, representing measurement error or spatial variability at very short distances (the nugget); the point at which spatial correlation diminishes (the sill), and the distance at which the variogram reaches the sill, beyond which points are effectively uncorrelated (the range). A function is then typically fit to the empirical values by adjusting for the parameters of nugget, sill, and range, in order for a continuous model to be obtained that best fits the empirical distribution of the data. Such a model of how the variance varies with distance forms the basis for \emph{kriging}, a more sophisticated interpolation method that we employ in our experiments and describe next. In our experiments, we automatically fit the variogram by way of the best least squares fit to the data.

\subsubsection{Ordinary Kriging}
Kriging \citep{oliver1990kriging} uses the variogram to calculate weights that account for both distance and spatial correlation, providing more accurate estimates at unsampled locations. In ordinary kriging, the value at an unknown location \( x_0 \), denoted as \( \hat{Z}(x_0) \), is a weighted sum of the known values \( Z(x_i) \) at nearby locations:

\[
\hat{Z}(x_0) = \sum_{i=1}^{N} \lambda_i Z(x_i)
\]

The weights \( \lambda_i \) are determined using the variogram, giving more importance to closer points with stronger spatial correlation. To ensure the estimate is unbiased, the weights are constrained to sum to 1. The kriging weights are found by solving a system of equations based on the variogram, where a Lagrange multiplier enforces the constraint above. The weights are then applied to the known values to predict values at sites unseen in the data.

However, a key assumption behind ordinary kriging is that the spatial process which generates the values is stationary, where the mean and variance are assumed to be constant across space. Where this assumption does not hold, it may be necessary to incorporate auxiliary variables that help explain trends in the data by way of regression, which then allows kriging to be done on the residuals. 

\subsubsection{Regression Kriging} \label{RK}
Regression kriging \citep{hengl2007regression} extends ordinary kriging by incorporating auxiliary variables to account for trends in the data that might otherwise violate the stationarity assumption. These auxiliary variables are commonly assumed to have a linear relationship with the variable of interest.

In regression kriging, the observed values \( Z(x_i) \) at known locations are assumed to follow a model of the form:

\[
Z(x_i) = m(x_i) + \epsilon(x_i),
\]

where \( m(x_i) \) is the drift term, which represents the trend at location \( x_i \), and \( \epsilon(x_i) \) is a spatially correlated random error with a mean of zero. The drift term is typically modeled as a linear combination of one or more auxiliary variables \( Y_j(x_i) \) at each location:

\[
m(x_i) = \sum_{j=1}^{p} \beta_j Y_j(x_i),
\]

where \( \beta_j \) are the regression coefficients, and \( Y_j(x_i) \) are the values of the auxiliary variables at location \( x_i \). The kriging weights \( \lambda_i \) are then computed by solving the kriging system, with the variogram used to model the spatial correlation of the residuals \( \epsilon(x_i) \).

In our experiments, we hypothesize that a correlation exists between speech-to-text BLEU and chrF2 scores and similarity to the standard variety (approximated by the highest performing variety). We therefore model similarity to the highest performing variety as the drift term, and compute the variogram on the basis of the residuals.



\subsection{Dialectometric Analysis}
To compare the geographical distribution of speech-to-text model performance against the geographical distribution of dialect similarity, we follow established approaches in dialectometry \citep{wieling2015advances} for both distance computation and visualization. Dialectometry \citep{nerbonne2010mapping} is a subfield of linguistics that aims to quantify dialect relations by way of quantitative techniques, which often employs edit distance on word-level data for such a quantification. We compute the aggregate dialect distance pair-wise between sites, where each site has 20 semantically equivalent audio recordings of words, resulting in a site-by-site matrix that is amenable to our visualization method. We present the details below.

\subsubsection{Linguistic Distance}
For the quantification of linguistic distance between dialect varieties, we follow \citet{bartelds-wieling-2022-quantifying} in adopting self-supervised speech representations for the extraction of features on semantically equivalent words, upon which dynamic time warping (DTW) can be applied for a measure of phonetic distance. Formally, let \( X \) and \( Y \) denote two lists of semantically equivalent words from two dialect varieties, where \( X = [w_X^1, w_X^2, \dots, w_X^n] \) and \( Y = [w_Y^1, w_Y^2, \dots, w_Y^n] \), with \( w_X^i \) and \( w_Y^i \) representing semantically equivalent items. For each pair of items \( (w_X^i, w_Y^i) \), we extract their acoustic features using XLSR-53, a self-supervised speech representation model, where we employed an off-the-shelf model finetuned on Italian Common Voice \citep{ardila2019common} data.\footnote{\url{https://huggingface.co/jonatasgrosman/wav2vec2-large-xlsr-53-italian}} Let \(f(w_X^i) \) and \( f(w_Y^i) \) represent the feature vectors for the words \( w_X^i \) and \( w_Y^i \), respectively.\footnote{We use the last hidden state in our experiments.} The acoustic distance \( dist(f(w_X^i), f(w_Y^i)) \) for the \( i \)-th pair is then computed by way of dynamic time warping, which obtains the distance between two time series that may vary in speed and length by computing the shortest path in a cost matrix. Following \citet{bartelds-wieling-2022-quantifying}, the distance is normalized by dividing over the length of the shortest path for a fair comparison between sites. The acoustic distance \( Dist(X, Y) \) between dialect varieties \( X \) and \( Y \) is then computed by averaging the pairwise distances across all \( n \) items in the lists:

\[
Dist(X, Y) = \frac{1}{n} \sum_{i=1}^{n} dist(f(w_X^i), f(w_Y^i))
\]


\begin{table}
\small
\SetTblrInner{rowsep=0pt}
\centering
\begin{tblr}{lll}
\toprule
 & \textbf{Pearson} & \textbf{Spearman} \\
\midrule
\textbf{chrF2} & -0.50 & -0.58 \\
\midrule
\textbf{BLEU} & -0.54 & -0.50  \\
\bottomrule
\end{tblr}
\caption{Correlation of speech-to-text performance with linguistic similarity to highest performing site.}
\label{table:dia_corr}
\end{table}

We apply $Dist(\cdot)$ to all pairwise combinations of dialect sites, resulting in a symmetric site-by-site distance matrix that is amenable to our visualization method described below.

\subsubsection{Multidimensional Scaling}
Having obtained a site-by-site distance matrix, we employ a visualization method based on dimensionality reduction described in \citet{nerbonne2010mapping}. \citet{nerbonne2010mapping} shows that a map depicting dialect relations as a continuous surface can be achieved by leveraging classical multidimensional scaling (MDS),\footnote{We use the implementation in dialectR \citep{shim-nerbonne-2022-dialectr}.} where MDS is a dimensionality reduction method that takes as input a distance matrix and aims to project it to a lower-dimensional space while aiming to preserve the distances in the original high-dimensional space. Formally, given data points \( X = \{x_1, x_2, \dots, x_n\} \), let \( D = [d_{ij}] \) be the distance matrix, where \( d_{ij} \) represents the distance between \( x_i \) and \( x_j \). MDS seeks to minimize the following objective, termed the stress function:

\[
S(Y) = \sqrt{\frac{\sum_{i,j} (d_{ij} - \| y_i - y_j \|)^2}{\sum_{i,j} d_{ij}^2}}
\]

where \( \| y_i - y_j \| \) is the Euclidean distance between \( y_i \) and \( y_j \) in the lower-dimensional space, and \( S(Y) \) represents the stress, a measure of how well the configuration \( Y \) preserves the original distances.

\citet{nerbonne2010mapping} proposes to reduce the pairwise distance matrix between dialect varieties to 3 dimensions with such an approach, which can then be converted to RGB values respectively (i.e. one dimension converted to one color), which are then overlayed on a map. This allows for color mixtures that visually depict gradual and sharp transitions, with the limitation of losing some of the information in the original distance matrix due to the dimensionality reduction.

\section{Results}
\subsection{Speech-to-Text Evaluation}

\begin{table}
\SetTblrInner{rowsep=0pt}
\scriptsize
\centering
\resizebox{\columnwidth}{!}{
\begin{tblr}{lrrrrrr}
\toprule
 & \textbf{N} & \textbf{Min} & \textbf{1st Q} & \textbf{Median} &  \textbf{3rd Q} & \textbf{Max} \\
\midrule
\midrule
\textbf{All Varieties} & 223 & 17.97 & 27.50 & 35.52 & 42.51 & 63.25 \\
\midrule
\midrule
\textbf{Tuscan} & 8 & \textbf{51.99} & \textbf{57.73} & \textbf{62.32} & \textbf{64.24} & \textbf{72.45} \\
\midrule
\textbf{Umbrian} & 13 & 48.03 & 49.67 & 54.80 & 60.82 & 70.45 \\
\midrule
\textbf{Abruzzian} & 8 & 34.10 & 38.00 & 42.89 & 46.86 & 53.16 \\
\midrule
\textbf{Venetian} & 31 & 31.53 & 37.86 & 41.76 & 46.50 & 57.72 \\
\midrule
\textbf{Sicilian} & 12 & 34.60 & 38.17 & 41.32 & 43.83 & 45.56 \\
\midrule
\textbf{Ligurian} & 15 & 26.88 & 32.73 & 37.03 & 39.77 & 48.54 \\
\midrule
\textbf{Trentinian} & 11 & 24.28 & 31.29 & 35.52 & 41.77 & 48.23\\
\midrule
\textbf{Lucanian} & 10 & 23.35 & 28.73 & 35.41 & 42.85 & 45.12\\
\midrule
\textbf{Molisan} & 15 & 25.95 & 30.94 & 34.26 & 39.70 & 43.26 \\
\midrule
\textbf{Apulian} & 6 & 21.08 & 24.30 & 32.48 & 38.99 & 42.20\\
\midrule
\textbf{Friulian} & 16 & 22.48 & 29.40 & 31.12 & 37.56 & 40.81\\
\midrule
\textbf{Ladin} & 9 & 17.97 & 21.72 & 27.52 & 28.71 & 38.10\\
\midrule
\textbf{Piedmontese} & 13 & 19.93 & 25.15 & 26.40 & 28.68 & 30.34\\
\midrule
\textbf{Lombardian} &  23 & 18.50 & 23.50 & 25.84 & 30.24 & 39.07\\
\midrule
\textbf{Sardinian} & 14 & 20.04 & 22.44 & 23.69 & 25.13 & 27.50 \\
\bottomrule
\end{tblr}
}
\caption{Statistical summary of chrF2 scores for all Italian dialects and for dialect groupings with more than 5 sites, in decreasing order by median.}
\label{table:s2t_summary_chrf}
\end{table}

\begin{table}
\SetTblrInner{rowsep=0pt}
\scriptsize
\centering
\resizebox{\columnwidth}{!}{
\begin{tblr}{lrrrrrr}
\toprule
 & \textbf{N} & \textbf{Min} & \textbf{1st Q} & \textbf{Median} &  \textbf{3rd Q} & \textbf{Max} \\
\midrule
\midrule
\textbf{All Varieties} & 223 & 0.00 & 4.19 & 7.98 & 14.50 & 29.67 \\
\midrule
\midrule
\textbf{Tuscan} & 8 & \textbf{14.75} & \textbf{20.19} & \textbf{32.72} & \textbf{39.18} & \textbf{49.72} \\
\midrule
\textbf{Umbrian} & 13 & 12.53 & 19.45 & 27.56 & 34.32 & 46.08 \\
\midrule
\textbf{Abruzzian} & 8 & 7.58 & 12.80 & 16.14 & 19.16 & 27.69 \\
\midrule
\textbf{Lucanian} & 10 & 3.48 & 4.31 & 11.65 & 14.42 & 16.75 \\
\midrule
\textbf{Sicilian} & 12 & 3.45 & 6.63 & 11.43 & 13.70 & 21.19\\
\midrule
\textbf{Venetian} & 31 & 3.31 & 7.75 & 10.53 & 18.28 & 31.06 \\
\midrule
\textbf{Molisan} & 15 & 2.20 & 6.39 & 8.97 & 12.50 & 21.10 \\
\midrule
\textbf{Apulian} & 6 & 1.53 & 2.24 & 8.65 & 13.14 & 18.30 \\
\midrule
\textbf{Ligurian} & 15 & 2.65 & 4.27 & 7.45 & 10.20 & 17.97 \\
\midrule
\textbf{Trentinian} & 11 & 3.75 & 4.38 & 7.05 & 10.96 & 16.51 \\
\midrule
\textbf{Friulian} & 16 & 1.75 & 3.02 & 6.62 & 11.41 & 16.96 \\
\midrule
\textbf{Lombardian} & 23 & 1.62 & 2.42 & 4.79 & 7.43 & 11.45 \\
\midrule
\textbf{Ladin} & 9 & 1.36 & 1.80 & 4.35 & 7.04 & 11.57 \\
\midrule
\textbf{Sardinian} & 14 & 1.99 & 2.34 & 4.30 & 6.12 & 8.85 \\
\midrule
\textbf{Piedmontese} & 13 & 1.74 & 1.92 & 3.78 & 5.90 & 9.21 \\
\bottomrule
\end{tblr}
}
\caption{Statistical summary of BLEU scores for all Italian dialects and for dialect groupings with more than 5 sites, in decreasing order by median.}
\label{table:s2t_summary_bleu}
\end{table}

\begin{figure*}[htp]
  \centering
  \subfigure{\includegraphics[width=0.375\linewidth]{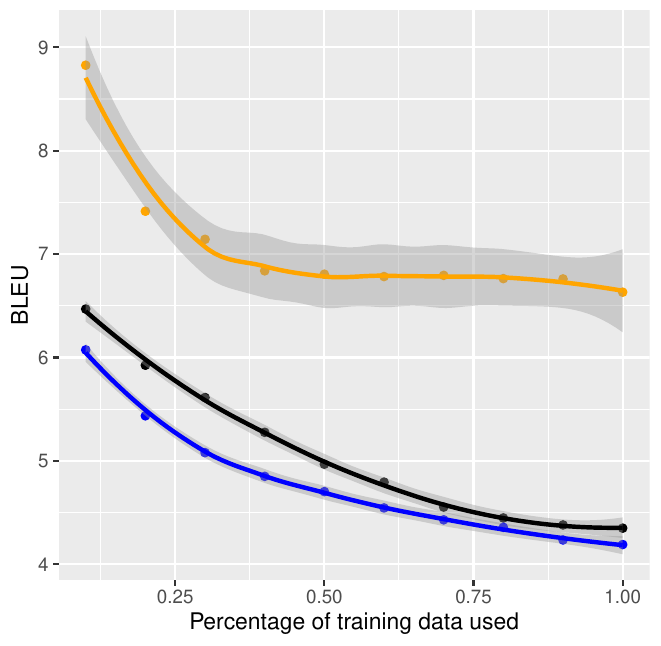}}\quad
  \subfigure{\includegraphics[width=0.375\linewidth]{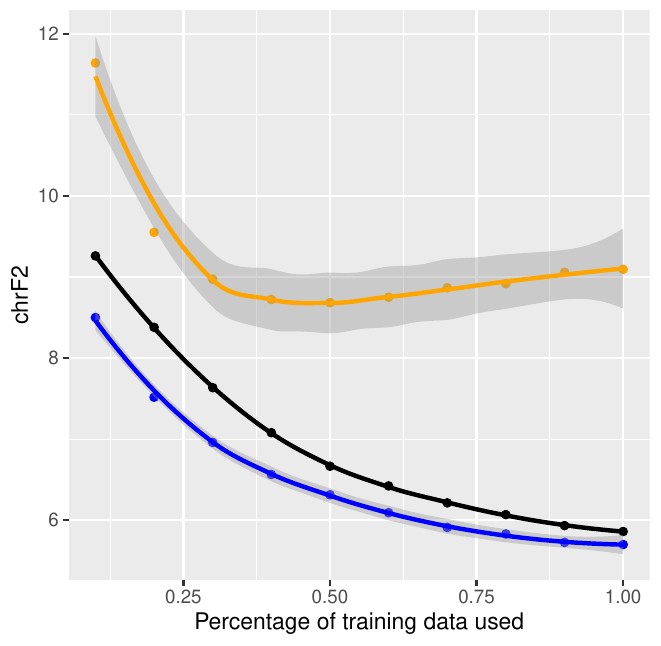}}
  \caption{Effect of training data size on geostatistical interpolation. Orange is nearest neighbor interpolation (NN); black is inverse distance weighting (IDW); blue is regression kriging (RK).}
  \label{fig:geocurve}
\end{figure*}

The chrF results are geographically interpolated and plotted in \autoref{fig:maps}. We observe in the map that zero-shot performance is particularly high in the Tuscan region, where the highest performing dialect site also resides, as shown in \autoref{table:s2t_summary_chrf} and \autoref{table:s2t_summary_bleu}. Prior literature has established standard Italian to be modelled after Tuscan varieties \citep{hall1980italian, wieling2014lexical}, which corroborates the trend observed in our results. The similarity of Umbrian dialects with Tuscan dialects is also observable in Umbrian dialects ranking second after Tuscan in \autoref{table:s2t_summary_chrf} and \autoref{table:s2t_summary_bleu}. Furthermore, \autoref{table:dia_corr} details the correlation between linguistic similarity and the performance scores on the level of dialect sites, where the Pearson correlation for the chrF score is -0.50 and for BLEU -0.54, and the Spearman correlation is -0.58 for chrF and -0.50 for BLEU, suggesting---to answer RQ1---a strong correlation between similarity to the standard (as approximated by the highest performing site) and speech-to-text performance.

\subsection{Dialectometric Analysis}
In \autoref{fig:maps}, colors that are more similar in the dialectometry map indicate more linguistic similarity. We observe the green-tinted areas such as Tuscany correspond with higher-performing regions, which we interpret to be similarity to the standard variety. The greenness in north Sardinia is documented in prior literature \citep{cugno2022italian}, where the dialects spoken there---Gallurese and Sassarese---are considered to be Southern Corsican varieties, where Corsican is considered to be strongly influenced by Tuscan. Similarly, the green tints in Sicily correspond with observations made of the dataset in prior literature \citep{la-quatra-etal-2024-speech}, where it is noted that varieties such as Sicilian contain a considerable amount of standard Italian presence in the data, suggesting some samples to exhibit language mixing between standard and dialect.

\subsection{Geostatistical interpolation}
Building on the observation that there is a clear geographical signal in both the performance and dialectometry maps, we next turn to RQ2 and measure to what extent the incorporation of geographical knowledge helps predict zero-shot speech-to-text performance at unseen sites. 


\begin{table}[h!]
\SetTblrInner{rowsep=0pt}
\centering
\begin{tblr}{llll}
\toprule
 & \textbf{NN} & \textbf{IDW} & \textbf{RK} \\
 \midrule
\textbf{chrF2} & 9.09 & 5.86 & \textbf{5.70} \\
\midrule
\textbf{BLEU} & 6.63 & 4.35 & \textbf{4.19} \\
\bottomrule
\end{tblr}
\caption{RMSE of geostatistical interpolation methods on unseen sites. NN: Nearest Neighbor, IDW: Inverse Distance Weighting, RK: Regression Kriging.}
\label{table:geo_rmse}
\end{table}

As shown in \autoref{table:geo_rmse}, geostatistical interpolation is highly predictive of both BLEU and chrF scores, with RMSE scores going to as low as 4.19 and 5.70 by regression kriging, our best method. \autoref{fig:geocurve} additionally show the effect of training data size on the regression prediction performance. In all results, we observe that the incorporation of distance and covariance between samples as weighting improves over an interpolation by only the value of the nearest neighbor, where regression kriging is consistently the best performing method, followed by inverse distance weighting.

\section{Discussion}
\subsection{Evaluating Dialects as a Continuum}
In \autoref{fig:maps}, we observe that for Italian dialects, the evaluation results stand in a continuum that bears similarity to the map of dialect similarity relations. Furthermore, even when one restricts the examined samples to Tuscan varieties, we observe in \autoref{table:s2t_summary_chrf} and \autoref{table:s2t_summary_bleu} that there is still a disparity that extends the further away one is from the highest performing sites. Our results highlight the need for work in dialect evaluation to take into account the continuous nature of dialects that may exhibit even within dialects often perceived as falling under the same ``category''. 

While prior work \citep{kantharuban-etal-2023-quantifying} has made important headway in individually evaluating non-standard varieties as separate categories and highlighting the performance disparity when compared against the standard, regional variation also exhibits within non-standard varieties such as AAVE \citep{jones-2015-aave-variation}. Our results suggest that performance prediction in AAVE would arguably also exhibit regional differences, potentially patterning based on how similar the regional varieties are to standard English (e.g. in urban sites), although empirical work is needed to confirm this hypothesis. Importantly, our findings highlight that an evaluation of dialects that is insufficiently balanced geographically therefore carries the risk of overly optimistic views towards model performance at geographically marginal sites, which in turn may lead to social harm towards the subgroups which speak it.

\subsection{Geo-Based Performance Prediction}
Extending on the claim of \citet{ahuja-etal-2022-multi}, who highlight the role of geography in predicting how well the pivot language for finetuning generalizes to the target language, we explicitly utilize geostatistical methods in our work on geographically proximate dialects. We find both the distance between dialect sites and the covariance between sites to be useful for predicting zero-shot speech-to-text performance at unseen sites. Our results emphasize the geographical structure of dialects, and point to the possibility of leveraging such geographical structure for multilingual transfer between dialects.

\section{Related Work}
\citet{kantharuban-etal-2023-quantifying} stands as the work most similar to our own, where LLMs for both speech and for text are evaluated on the tasks of Machine Translation and Automatic Speech Recognition on regional dialects of high and low-resource languages. They find model performance on such varieties to be highly correlated with lexical and phonetic similarity to the highest performing variety. Their work however considers dialect varieties \textit{categorically} by considering regional varieties such as Argentinian and Chilean Spanish under the same language of Spanish. Our work instead proposes to work on varieties on a continuous basis across geographically nearby varieties.


With regard to treating varieties as a continuum, \citet{grieve2024sociolinguisticfoundationslanguagemodeling} hypothesize that language models inherently model varieties of language, and propose to leverage sociolinguistic expertise to identify underrepresented varieties. \citet{bafna-etal-2024-evaluating} model performance degradation of LLMs on closely-related varieties by way of synthetic noise in different linguistic dimensions. In dialectology, the notion of dialects as a continuum is well-established, where computational work in quantifying such a continuum abounds in the field of dialectometry \citep{heeringa2001dialect, wieling2015advances, nerbonne2010mapping}.

With regard to bias against dialect varieties in ASR, 
\citet{feng2021quantifying} show that Dutch ASR systems perform worse on Flemish speakers compared to speakers of all regions in the Netherlands. \citet{kulkarni2024balancing} show different ASR systems to exhibit different performance biases across 11 states in Brazil, tested on scripted speech and with categorical boundaries between states. \citet{chang2024self} document how self-supervised representations still exhibit a performance disparity in ASR upon AAVE. Our work shows that there is geographical structure in dialect speech-to-text bias that is continuous and correlated with social variables, enabling more fine-grained studies on speech-to-text bias in other languages.

\section{Conclusion}
In this paper, we 
conceptualize dialect relations as a continuum. 
We find zero-shot performance of speech-to-text systems on dialects to pattern similarly to a measure of similarity to the standard variety, observing strong correlations. We cross-examine our results against established research in linguistic dialectometry. Furthermore, we introduce geostatistical methods that are predictive of zero-shot performance at held-out sites. 
Our work highlights the need for more research on non-standard varieties that takes into account the continuum nature of dialects. Doing so holds the potential for uncovering bias against speakers of non-standard varieties and helps  work toward closing prior evaluation gaps. 

\section{Limitations}
Our study focuses on related varieties of Italian dialects, which potentially limits the generalizability of our findings. Furthermore, we approximate the standard variety by the best performing dialect variety, which may affect the correlation results depending on how similar the best performing Tuscan variety actually is to standard Italian.
Future work should examine how well our insights generalize to other dialect continua, where dialect corpora similar to Vivaldi exist for Sino-Tibetan \citep{zhongguoyuyan} and Alpine varieties \citep{rabanus2023alpilink}. Similarly, whether a continuum-based disparity may likewise also exhibit in text-based regional linguistic data remains an open question that future work can explore.

\section{Acknowledgements}
We would like to thank the members of the MaiNLP lab and the anonymous reviewers for their invaluable feedback. The icons featured in \autoref{fig:maps} are from Freepik (Italy, neural network)\footnote{\url{https://www.flaticon.com}} and SVG Repo (pin, arrow, sound wave).\footnote{\url{https://www.svgrepo.com}} Lastly, we recognize the support for SS and BP through the ERC Consolidator Grant 101043235.

\bibliography{custom, anthology}

\end{document}